\documentclass[conference]{IEEEtran}
\IEEEoverridecommandlockouts
\usepackage{graphicx}
\usepackage{multirow}
\usepackage{booktabs}
\usepackage[table]{xcolor}
\usepackage{cite}
\usepackage{amsmath,amssymb,amsfonts}
\usepackage{algorithmic}
\usepackage{graphicx}
\usepackage{textcomp}
\usepackage{xcolor}
\DeclareRobustCommand*{\IEEEauthorrefmark}[1]{\raisebox{0pt}[0pt][0pt]{\textsuperscript{\footnotesize\ensuremath{#1}}}}
\begin{document}

\title{Breaking Shallow Limits: Task-Driven Pixel Fusion for Gap-free RGBT Tracking
}

\author{
	\IEEEauthorblockN{
		Andong Lu\IEEEauthorrefmark{1}, 
        Yuanzhi Guo\IEEEauthorrefmark{2},
		Wanyu Wang\IEEEauthorrefmark{2}, 
		Chenglong Li\IEEEauthorrefmark{2}*, 
		Jin Tang\IEEEauthorrefmark{1} 
		and Bin Luo\IEEEauthorrefmark{1}} 
	\IEEEauthorblockA{\IEEEauthorrefmark{1}School of Computer Science and Technology, Anhui University}
	\IEEEauthorblockA{\IEEEauthorrefmark{2}School of Artificial Intelligence, Anhui University\\  adlu\_ah@foxmail.com}
}


 \maketitle

\begin{abstract}
Current RGBT tracking methods often overlook the impact of fusion location on mitigating modality gap, which is key factor to effective tracking. Our analysis reveals that shallower fusion yields smaller distribution gap. However, the limited discriminative power of shallow networks hard to distinguish task-relevant information from noise, limiting the potential of pixel-level fusion.
To break shallow limits, we propose a novel \textbf{T}ask-driven \textbf{P}ixel-level \textbf{F}usion network, named \textbf{TPF}, which unveils the power of pixel-level fusion in RGBT tracking through a progressive learning framework.
In particular, we design a lightweight Pixel-level Fusion Adapter (PFA) that exploits Mamba’s linear complexity to ensure real-time, low-latency RGBT tracking. To enhance the fusion capabilities of the PFA, our task-driven progressive learning framework first utilizes adaptive multi-expert distillation to inherits fusion knowledge from state-of-the-art image fusion models, establishing robust initialization, and then employs a decoupled representation learning scheme to achieve task-relevant information fusion.
Moreover, to overcome appearance variations between the initial template and search frames, we presents a nearest-neighbor dynamic template updating scheme, which selects the most reliable frame closest to the current search frame as the dynamic template. 
Extensive experiments demonstrate that TPF significantly outperforms existing most of advanced trackers on four public RGBT tracking datasets. The code will be released upon acceptance.
\end{abstract}

\section{Introduction}
\label{sec:intro}

Visual tracking is a fundamental task in computer vision that aims to localize a target in subsequent frames given its initial state. 
Since the complementary advantages of RGB and thermal modalities can enhance tracking robustness under challenging conditions, RGB-Thermal (RGBT) tracking receives great attention from researchers in recent years, and extensive works~\cite{Zhang_CVPR22_VTUAV,lu2024breaking, zhang2023efficient,BAT2024,SDSTrack,OneTracker} show impressive tracking performance. 
%

\begin{figure*}[t]
    \centering
    \includegraphics[width=1.0\linewidth]{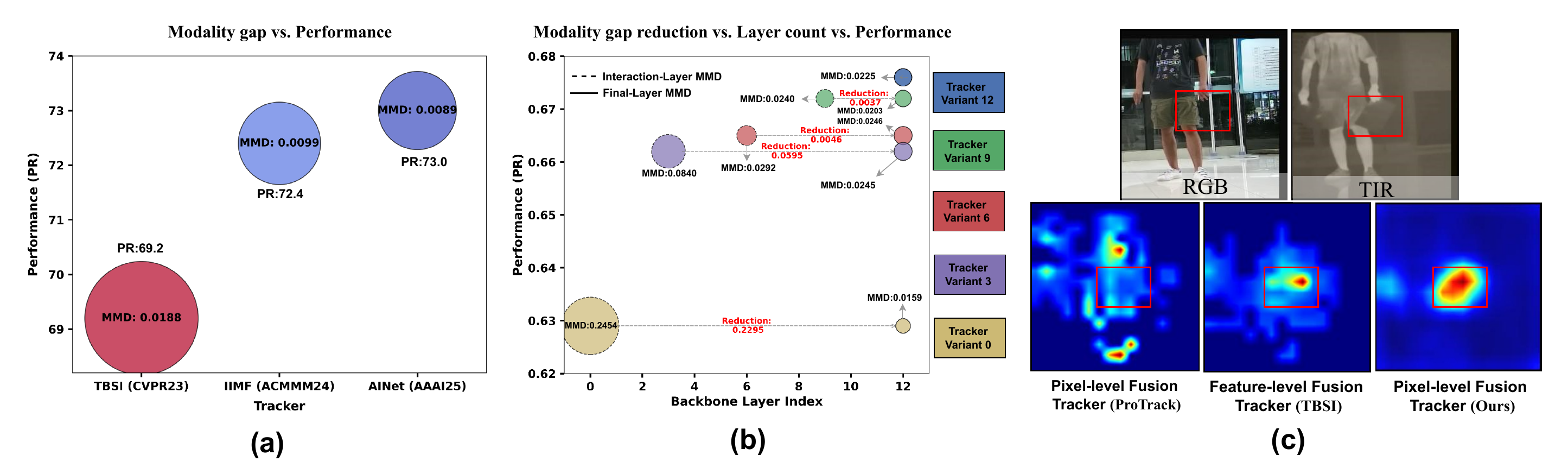}
     \caption{Analysis of key contradictions in RGBT tracking. (a) Modality gap vs. performance: Comparison of the modality gap at the final backbone layer across trackers with different performance levels. (b) Modality gap vs. interaction location: Comparison of five RGBT variants~\cite{ostrack} with identical interaction modules~\cite{TBSI} but different interaction locations (0/3/6/9/12), analyzing the modality gap at the interaction and final layers, along with tracking performance.  (c) Visual comparison of fusion methods: Fusion features at the final layer for pixel-level fusion, feature-level fusion, and the proposed method. MMD values are computed on the entire LasHeR~\cite{li2021lasher} test set.
     }
    \label{motivation_fig}
\end{figure*}

A critical challenge in RGBT tracking lies in designing effective fusion strategies to exploit cross-modal information. Existing approaches primarily follow two paradigms:
Feature-level fusion~\cite{TBSI,BAT2024,ViPT,QAT2023,chen2024simplifying,lu2024rgbt} performs modality interaction in deep network layers, achieving high accuracy but requiring multiple feature extractions, which incurs high computational cost. For example, Chen~\emph{et al.}~\cite{chen2024simplifying} integrate a cross-attention based modality-sharing aggregation module after the fourth block of the ViT backbone, effectively merging information, while Lu~\emph{et al.}~\cite{lu2024rgbt} propose a progressive fusion Mamba to interact features across all layers for robust tracking.
In contrast, Pixel-level fusion\cite{ProRGBTTrack,Exploring_IF_2023} merges raw inputs at shallow layers, significantly improving efficiency but suffering from performance degradation. For instance, Yang~\emph{et al.}~\cite{ProRGBTTrack} employ pixel-level fusion based on a prompt learning in multimodal tracking, while Tang \emph{et al.}~\cite{Exploring_IF_2023} introduce an existing image fusion network for RGBT tracking.  
This contrast reveals a trade-off: shallow interaction improves efficiency but sacrifices accuracy, while deep interaction ensures performance at high computational cost.

Our preliminary investigation yields two key insights that motivate this study. 
First, high-performance feature-level interaction trackers exhibit a smaller modality gap (quantified by Maximum Mean Discrepancy~\cite{gretton2012kernel}, MMD) in their final backbone layers, suggesting that improved tracking accuracy is closely correlated with a minimized modality gap, as illustrated in Figure~\ref{motivation_fig}(a). 
Second, contrary to intuition, shallow interaction achieves lower final-layer MMD than deep interaction, yet yield the poorest tracking accuracy, as shown in Figure~\ref{motivation_fig}(b). This inconsistency suggests that earlier interactions allow more backbone layers to align modalities, but may inadvertently harm feature discriminability.
To diagnose this issue, we visualize the fusion feature maps of both fusion paradigms. As shown in Figure~\ref{motivation_fig}(c), pixel-level fusion tracker~\cite{ProRGBTTrack} responds strongly to both targets and distractors, whereas feature-level fusion tracker~\cite{TBSI} effectively suppresses background interference. This suggests that shallow interactions lack sufficient discriminative power.
In essence, although shallow interactions reduce the modality gap, their limited discriminative power ultimately impairs robust tracking performance, motivating our search for a more effective pixel-level fusion strategy.

To break shallow limits, we propose a novel task-driven pixel-level fusion network, which achieves efficient and high-performance RGBT tracking through a progressive learning framework.
We first adopt Mamba with linear computational complexity to build a lightweight \textbf{P}ixel-level \textbf{F}usion \textbf{A}dapter (PFA) that meets real-time tracking requirements. With only 14.3KB of parameters, PFA achieves an impressive image fusion efficiency of 201.8 FPS.
Next, we introduce the \textbf{T}ask-driven \textbf{P}rogressive \textbf{L}earning (TPL) framework to address the challenge of insufficient discriminative ability in shallow networks.  
In the first phase, multi-expert adaptive distillation allows PFA to inherit strengths from multiple state-of-the-art fusion models~\cite{zhao2023cddfuse,yi2024diff,li2024mambadfuse}, providing a robust initialization. However, shallow networks tend to retain task-irrelevant information due to their limited discriminative power.

To address this, we introduce decoupled representation fine-tuning in the second phase. This strategy separates RGB and TIR features into task-relevant and task-irrelevant components, enabling the pixel-level fusion tracker to focus on the relevant features. Specifically, two auxiliary branches independently encode task-irrelevant information from each modality, while the tracking branch extracts task-relevant features by combining the modalities. This process is guided by three key losses, including a task loss to ensure relevance, a repulsion loss to separate task-irrelevant information, and a reconstruction loss to maintain input completeness.  This ensures that pixel-level fusion tracker focus on integrating task-relevant features while filtering out irrelevant data.
In addition, appearance changes of targets between the initial template and search frames over time introduce additional challenges for pixel-level fusion. To overcome this issue, we design a simple yet effective Nearest-Neighbor Dynamic Template Update (NDTU) strategy to select the closest reliable tracking result to the current search region as a dynamic template.
In summary, our major contributions are as follows:
\begin{itemize}
    \item We introduce a novel pixel-level fusion strategy for RGBT tracking that explicitly addresses the modality gap by strategically leveraging fusion location demonstrating for the first time how early fusion can better align modalities for enhanced performance.
    
    \item We develop a lightweight Pixel-level Fusion Adapter (PFA) based on Mamba's linear complexity, achieving real-time tracking with minimal parameters.
    
    \item We propose a task-driven progressive learning framework that combines multi-expert adaptive distillation and decoupled representation fine-tuning to isolate task-relevant features from distracting information.

    \item Extensive experiments on four RGBT tracking benchmarks show that our method achieves advanced performance against existing methods and faster tracking speed. 

\end{itemize}

\section{Related Work}

\subsection{RGBT Tracking}
RGBT tracking has seen significant advancements by integrating RGB and TIR data to achieve robust tracking across diverse environments. Existing methods primarily focus on feature-level fusion, while pixel-level fusion has not been sufficiently studied.
Feature-level fusion approaches~\cite{DMCNet2022,TBSI,ViPT,BAT2024,SDSTrack} are commonly designed to incorporate modality interactions within the feature extractor. For instance, Hui~\emph{et al.}~\cite{TBSI} propose to use fusion templates to guide multi-layer interactions between modalities in specific layers. Cao~\emph{et al.}~\cite{BAT2024} propose bi-directional adapters to perform RGB and TIR feature interactions between all layers. In addition, several studies~\cite{2021MANet++,QAT2023,li2019manet} only incorporate two modality features in the last layer by enhancing modal-specific representation. 
Despite these advances, the feature-level fusion method struggle to balance efficiency and fully exploiting the backbone network to mitigate modality differences.
Due to the straightforward of pixel-level fusion, several studies~\cite{zhang2019multi,tang2023exploring,ProRGBTTrack,ding2023x} also try to explore the application of pixel-level fusion in RGBT tracking. For example, Tang~\emph{et al.}~\cite{tang2023exploring} introduces the knowledge of image fusion applied to the pixel-level in RGBT tracking. Yang~\emph{et al.}~\cite{ProRGBTTrack} achieves pixel-level fusion between modalities and achieves efficient tracking through simple addition. However, existing pixel-level fusion methods are not comparable to feature-level fusion methods in terms of performance, which limits their further development.

\begin{figure*}[h]
	\centering
	\includegraphics[width=0.92\linewidth]{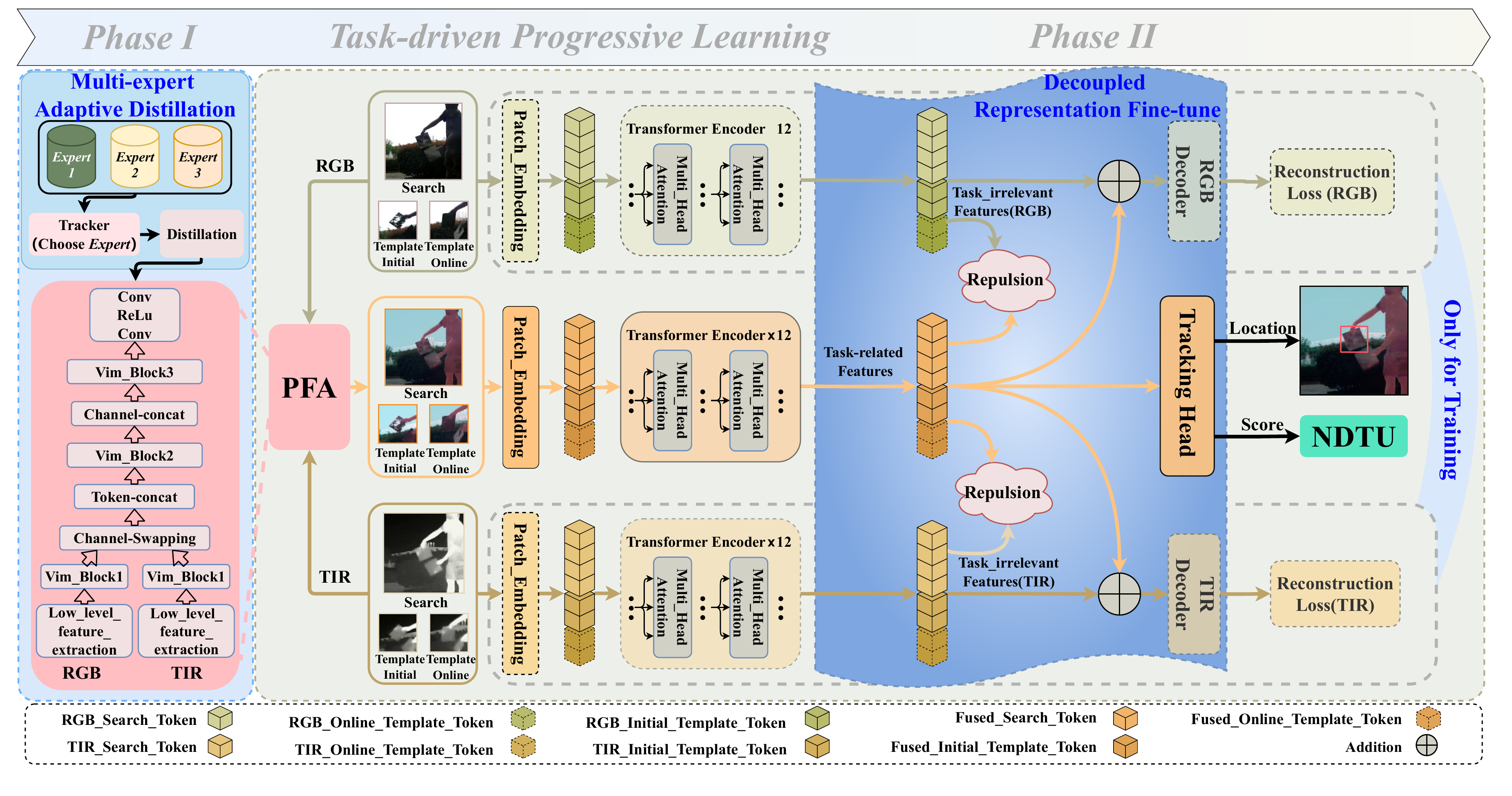}
	\caption{
		Overall framework of Task-driven Pixel-level Fusion (TPF) network for RGBT tracking.
	}
	\label{fig:overall_framework}
\end{figure*}

\subsection{Visible Infrared Image fusion}
Recent advances~\cite{huang2022reconet,huang2023learning,zhang2023ingredient,zhao2023cddfuse,yi2024diff} in deep learning-based image fusion focus heavily on achieving high-quality fusion results, often by refining network structures and loss functions. 
For instance, CDDFuse~\cite{zhao2023cddfuse} upgraded the decomposition network to a Transformer-CNN dual-stream structure. DDFM~\cite{yi2024diff}, for the first time, utilized denoising diffusion models for image fusion tasks.
However, these methods mainly emphasize fusion itself and lack consideration of downstream tasks.
Although some efforts~\cite{li2023lrrnet,liu2023multi,liu2022target} combine image fusion with specific downstream tasks through a cascading framework for task-oriented fusion to improve performance. 
However, most methods simply apply multi-task loss to optimize the overall model, which limits the fusion network to adaptively refine feature representations for subsequent tasks.

\section{Method}

\subsection{Framework Overview} 
\label{3.1}
We provide a detailed description of the proposed framework (TPF), and its pipeline is illustrated in Figure~\ref{fig:overall_framework}. 
During training, TPF comprises a pixel-level fusion adapter (PFA), three branches (including a RGB branch, a TIR branch, a fusion branch), and a tracking head. 
In testing, only the PFA, fusion branch, and tracking head are used.

In specific, for the given RGB and TIR modality frames, the search and template frames are first send to PFA for obtaining fused search and template frames, respectively.
Next, these the original two modality frames and fused frames are partitioned into patches with the size of $p\times p$ using a learnable patch embedding layer, and flattened to obtain search token sequences ($S_{rgb}$, $S_{tir}$, $S_{f}$), and four template token sequences ($T_{rgb}$, $T_{tir}$, $T_{f}$), respectively.
Following~\cite{ostrack}, we also add learnable position embeddings to the above tokens to provide positional prior information. 
Then, we concatenate search and template frame token sequences for the three groups, denoted as $I_{rgb} = [S_{rgb}, T_{rgb}]$, $I_{tir} = [S_{tir}, T_{tir}]$, and $I_{f} = [S_{f}, T_{f}]$. We feed $I_{rgb}$ and $I_{tir}$ into the RGB and TIR reconstruction branches, respectively, while $I_{f}$ send into the fusion branch. 
There two reconstruction branches consisting of a standard Transformer encoder~\cite{ViT_network} to extract modality-specific information and a decoder to reconstruct original input, while the fusion branch contains a standard Transformer encoder to extract fused features from fused image and a tracking head for object localization and regression. Note that the encoders of these branches share the same architecture but have independent parameters, and RGB and TIR reconstruction branches are only used during training. 

\subsection{Pixel-level Fusion Adapter}
\label{3.2}
To achieve pixel-level fusion, we do not directly concatenate or sum image matrices as in other methods~\cite{zhang2019multi,ProRGBTTrack}. Drawing inspiration from visible and infrared image fusion studies~\cite{zhao2023cddfuse,liu2023multi}, we design a lightweight pixel-level fusion adapter (PFA). 
To robustly aggregate cross-modal information while meeting real-time tracking requirements, we leverage the Mamba architecture with global modeling capabilities and linear computational complexity to achieve efficient pixel-level fusion between two modal images. We present the network details of the PFA in Figure~\ref{fig:overall_framework}.
%
In particular, the PFA consists primarily of three ViM blocks~\cite{vim} with channel dimensions of 8, 8, and 16, along with two convolutional layers.
First, each modality (RGB and TIR) is partitioned by a low-level feature extraction layer and then fed into separate ViM blocks to encode modality-specific features. To enhance cross-modality representation, we introduce a parameter-free channel swapping mechanism, addressing limitations in the ViM model’s channel dimension. Next, token and channel concatenation are applied to merge the two modalities along different feature dimensions, and two additional ViM blocks further encode this fused information. Final, we decode the fused features into images using a convolutional layer with efficient local detail modeling capability.

\subsubsection{Task-oriented Progressive Learning}
To balance the efficiency of pixel-level fusion with enhanced discriminative capability, we propose a task-oriented progressive learning (TPL) framework that consists of two key components: Multi-expert Adaptive Distillation and Decoupled Representation Fine-tuning.

\noindent\textbf{Multi-expert Adaptive Distillation.} PFA can provide efficient image fusion due to its low parameter settings, but also limits fusion performance. Although several methods~\cite{Exploring_IF_2023} attempt to employ distillation strategies to improve the fusion capability, it is hard to provide adequate guidance in a variable tracking scenario with only relying on the knowledge of a single expert model.
To overcome this limitation, we propose a method called Multi-expert Adaptive Distillation (MAD), which aims to inherit superior fusion capabilities from multiple image fusion models with different architectures. MAD introduces an additional single-stream tracker during the distillation phase to evaluate the performance of fused images from different experts in tracking tasks. Hence, MAD can adaptively adjust the importance weights of different experts, ensuring the selection of the optimal expert for distillation in various scenarios.

To be specific, we chose three different image fusion models, including a CNN-based model~\cite{zhao2023cddfuse}, a diffusion-based model~\cite{yi2024diff}, and a mamba-based model~\cite{li2024mambadfuse}, as the three experts of PFA. Given an RGB image and a TIR image, let $I_{PFA}$, $I_{c}$, $I_{d}$, $I_{m}$ denote the fused outputs from PFA, the CNN-based, the diffusion-based, and the Mamba-based models respectively. The adaptive weights are computed as:
\begin{align}  
\mathcal{W}_c, \mathcal{W}_d, \mathcal{W}_m = \frac{\mathcal{T}(I_c), \mathcal{T}(I_d), \mathcal{T}(I_m)}{\sum_{i \in \{c, d, m\}} \mathcal{T}(I_i)},
\label{eq:weight}
\end{align} 
\\
where $\mathcal{T}$ indicate the IoU (Intersection over Union) predictor of the auxiliary single-stream tracker, and $\mathcal{W}_i$ is the importance weight for the $i$-th expert. The distillation loss is then defined as:
\begin{align}  
\mathcal{L}_{dist} = \sum_{i \in \{c, d, m\}} \mathcal{W}_i \cdot (I_{PFA} - I_{i})^2.
\label{eq:distillation_loss}  
\end{align} 
Minimizing $\mathcal{L}_{dist}$ allows the PFA to inherit the fusion strengths of multiple experts while preserving its lightweight computational complexity.

\begin{figure}[ht]
	\centering
	\includegraphics[width=0.92\linewidth]{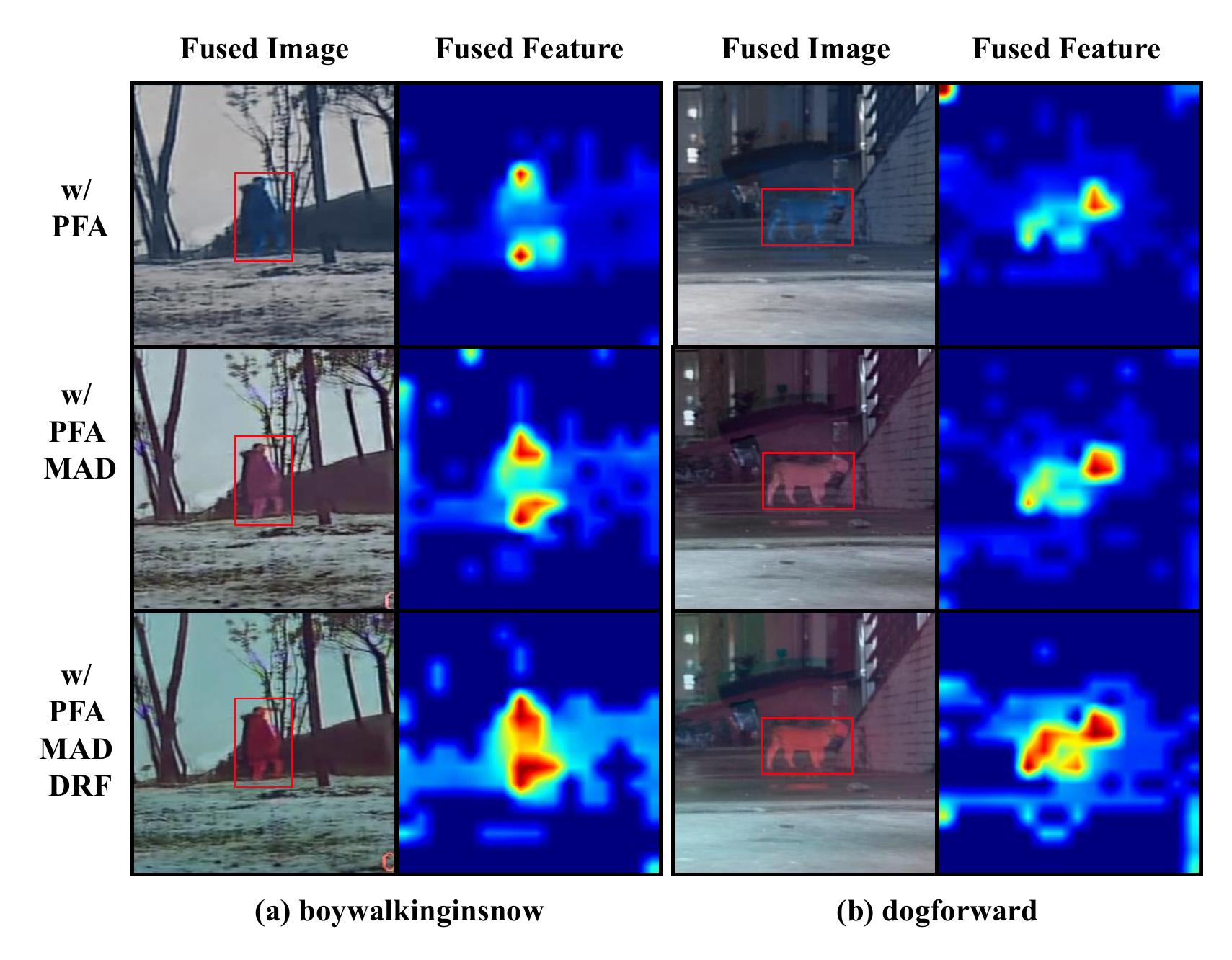}
	\caption{
	Visualization of fused images and fused features between different components.
	}
	\label{fig:vis_component}
\end{figure}

\noindent\textbf{Decoupling Representation Fine-tuning.} 
Since all expert models remain oriented towards human vision, further fine-tuning of the PFA is necessary for effective adaptation to the tracking task.
To this end, we propose a novel Decoupled Representation Fine-tuning (DRF) strategy to explicitly guide the PFA and tracker in integrating and extracting task-relevant information. 
As illustrated in Figure~\ref{fig:vis_component}, we present the fused images and features after the gradual introduction of each component in the TPF, and it can be seen that the response of the model to the target region is progressively enhanced in both images and features. This demonstrates the effectiveness of the proposed method.

In particular, we introduce two auxiliary branches, each consisting of a feature extraction backbone (mirroring the tracking branch structure) and a modality-specific decoder. 
These auxiliary branches operate in parallel with the tracking branch to encode RGB features $\mathcal{F}_{rgb}$, TIR feature $\mathcal{F}_{tir}$ and fused feature $\mathcal{F}_{f}$ from RGB, TIR and fused images, respectively. 
To decouple task-relevant and task-irrelevant features in each modality, we apply a task loss to optimize the learning of $\mathcal{F}_{f}$ and introduce a repulsion loss to push $\mathcal{F}_{rgb}$ and $\mathcal{F}_{tir}$ away from $F_{f}$. The repulsion loss is defined as:
\begin{align}  
\mathcal{L}_{rep} = \sum_{m \in \{rgb, tir\}} \left( \alpha - \frac{\mathcal{F}_f \cdot \mathcal{F}_m}{\| \mathcal{F}_f \| \| \mathcal{F}_m \|} \right)_+,
\label{eq:repulsion_loss}  
\end{align} 
where \((x)_+\) denotes \(\max(0, x)\), and \(\alpha\) is a margin parameter controlling the similarity threshold.
The repulsion loss prevents unnecessary overlap between these features during fine-tuning, allowing the model to focus on task-relevant information in each modality and minimizing interference.

To ensure that no crucial task information is lost during decoupling, we combine task-relevant and task-irrelevant features are fed into their corresponding decoders to reconstruct the original modality images, this process can be expressed as:
\begin{equation}
\mathcal{L}_{rec} = \sum_{m \in \{rgb, tir\}} \left\| \mathcal{D}_m \left( \mathcal{F}_m + \mathcal{F}_f \right) - I_m \right\|_2^2,
\end{equation}
where \( \mathcal{D}_{rgb} \) and \( \mathcal{D}_{tir} \) are the decoders for the RGB and TIR modalities, respectively, and $\left\| \cdot \right\|_2^2 $ denotes the mean squared error function. 
The reconstruction loss ensures that each modality retains both task-relevant and task-irrelevant information during fine-tuning, enabling dynamic modulation of these features and minimizing the loss of crucial task details during fusion.

Finally, we define the final loss during the fine-tuning phase as a combination of reconstruction loss, repulsion loss, and task-specific losses, formulated as follows:
\begin{equation}
\mathcal{L}_{drf} = \mathcal{L}_{task} +\lambda \times (\mathcal{L}_{rep} + \mathcal{L}_{rec}),
\end{equation}
where $\mathcal{L}_{\text{task}}$ represents the tracking task loss, as described in ~\cite{ostrack}. The coefficient $\lambda$ balances the contributions of each loss term, and setting it to 1 was optimal (see \textbf{Supplementary Material}).

In summary, the DRF strategy stabilizes the modality interaction learning process by explicitly separating task-relevant and task-irrelevant information, thereby enhancing fusion accuracy and robustness without compromising essential tracking details.


\subsection{Nearest-Neighbor Dynamic Template Updating}
\label{3.3}
Since the appearance differences between the initial template and search frames can change dramatically over time in pixel-level fusion scheme, it also limits the performance of the pixel-level fusion tracker.
To this end, a simple yet effective nearest-neighbor dynamic template updating (NDTU) strategy can be introduced to bridge the appearance differences by incorporating additional dynamic template.
To appropriately update the dynamic template during the tracking inference phase, we first identify reliable frame, defined as frame where the predicted target classification score $P$ exceeds a reliable threshold $p$. For all historically reliable frames, we only save the newest reliable frame $S_{new}$ for efficient. After every $N$ frames, we introduce $S_{new}$ as a dynamic template for the current frame. 
%
During the training phase, we introduce dynamically updated templates sampled from intermediate frames as extra input. 
Benefiting from these templates are fusion results of frames from different moments with different appearance differences, thus the model can better capture the appearance changes of the target by modeling the global relationships between the spatial and temporal dimensions.



\section{Experiments}

\begin{table*}[h]
	\centering
    \caption{PR/NPR and SR scores (\%) for advanced trackers on four datasets. 
 The best and second are the result of the $\color{red} red$ and $\color{blue} blue$.
 }
        \renewcommand\arraystretch{1.0}
	\resizebox{0.9\textwidth}{!}{
	\begin{tabular}{c|c|c|c|c|c|c|c|c|c|c|c|c|c}
            \toprule
         
          \multicolumn{2}{c|}{\multirow{2}{*}{\textbf{Methods}}} & \multirow{2}{*}{\textbf{Pub. Info.}} & \multirow{2}{*}{\textbf{Backbone}} & \multicolumn{2}{c|}{\textbf{GTOT}} & \multicolumn{2}{c|}{\textbf{RGBT210}} & \multicolumn{2}{c|}{\textbf{RGBT234}} & \multicolumn{3}{c|}{\textbf{LasHeR}}& {\textbf{FPS}} \\
          
		\multicolumn{-2}{c}{}& &&& \textbf{PR}$\uparrow$ & \textbf{SR}$\uparrow$ & \textbf{PR}$\uparrow$ & \textbf{SR}$\uparrow$ & \textbf{MPR}$\uparrow$ & \textbf{MSR}$\uparrow$ & \textbf{PR}$\uparrow$ & \textbf{NPR}$\uparrow$ & \textbf{SR}$\uparrow$ & $\uparrow$\\
        
		\midrule\midrule
            \multirow{22}{0.8cm}{\centering \small  \rotatebox{-90}{\textbf{\parbox{2cm}{ Feature$-$level\\ \hspace*{0.3cm}Fusion}}}}
		& DAPNet~\cite{zhu2019dense}        & ACM MM 2019 & VGG$-$M&88.2&70.7& $-$ & $-$& 76.6 & 53.7 &  43.1 & 38.3 & 31.4 & 2\\
		& CMPP~\cite{2020CMPP}              & CVPR 2020 & VGG$-$M & 92.6 & 73.8 & $-$ & $-$& 82.3 & 57.5 & $-$ & $-$& $-$ &1.3\\
		& CAT~\cite{2020CAT}                & ECCV 2020 & VGG$-$M &88.9&71.7& 79.2 & 53.3 & 80.4 & 56.1 & 45.0 & 39.5 & 31.4 & 20\\
		& ADRNet~\cite{ADRNet2021}          & IJCV 2021 & VGG$-$M&90.4&73.9& $-$ & $-$& 80.7 & 57.0 & $-$ & $-$& $-$& 25\\
		& JMMAC~\cite{2020JMMAC}            & TIP 2021 & VGG$-$M &90.2&73.2 & $-$ & $-$& 79.0 & 57.3& $-$ & $-$& $-$& 4\\
		& MANet++~\cite{2021MANet++}        & TIP 2021 & VGG$-$M&88.2&70.7& $-$ & $-$& 80.0 & 55.4 & 46.7 & 40.4 & 31.4 & 25.4 \\
		& APFNet~\cite{APFNet2022}          & AAAI 2022 & VGG$-$M&90.5&73.7& $-$ & $-$& 82.7 & 57.9 & 50.0 & 43.9 & 36.2 & 1.3\\
		& DMCNet~\cite{DMCNet2022}          & TNNLS 2022 & VGG$-$M &90.9&73.3& 79.7 & 55.5& 83.9 & 59.3 & 49.0 & 43.1 & 35.5&2.3 \\
            & HMFT~\cite{Zhang_CVPR22_VTUAV}    & CVPR 2022 & ResNet$-$50 &91.2&74.9& 78.6 & 53.5& 78.8 & 56.8 & $-$ & $-$& $-$& 30.2 \\
            & MFG~\cite{wang2022mfgnet}         & TMM 2022 & ResNet$-$18& 88.9 & 70.7 & 74.9 & 46.7 & 75.8 & 51.5 & $-$ & $-$& $-$& $-$ \\
            & DFNet~\cite{peng2022dynamic}      & TITS 2022 & VGG$-$M &88.1&71.9& $-$ & $-$& 77.2 & 51.3& $-$ & $-$& $-$& $-$ \\
            & MACFT~\cite{luo2023learning}      & Sensors 2023 & ViT$-$B&$-$&$-$&$-$&$-$& 85.7 & 62.2& 65.3 & $-$ & 51.4  &22 \\
            & CMD~\cite{zhang2023efficient}     & CVPR 2023 & ResNet$-$50&89.2&73.4&$-$& $-$& 82.4 & 58.4 & 59.0 & 54.6 & 46.4&  30 \\
            & ViPT~\cite{ViPT}                  & CVPR 2023 & ViT$-$B&$-$&$-$& $-$ & $-$& 83.5 & 61.7 & 65.1 & $-$ & 52.5& $-$ \\
            & TBSI~\cite{TBSI}                  & CVPR 2023 & ViT$-$B&$-$&$-$& 85.3 & 62.5& 87.1 & 63.7 & 69.2 &65.7 & 55.6 & 36.2 \\
            & QAT~\cite{QAT2023}                & ACM MM 2023 & ResNet$-$50&91.5&75.5& \color{blue}86.8 & 61.9&88.4 & 64.4  & 64.2 & 59.6 & 50.1 & 22 \\
            & BAT~\cite{BAT2024}                & AAAI 2024 & ViT$-$B&$-$&$-$& $-$ & $-$& 86.8 & 64.1 &70.2 & $-$ &56.3 & $-$ \\
            & TATrack~\cite{TATrack}            & AAAI 2024 & ViT$-$B&$-$&$-$& 85.3 & 61.8& 87.2 & 64.4  & 70.2 & $-$ & 56.1& 26.1 \\
            & OneTracker~\cite{OneTracker}      & CVPR 2024 & ViT$-$B&$-$&$-$& $-$ & $-$& 85.7 & 64.2 & 67.2 & $-$ & 53.8& $-$ \\
            & Un-Track~\cite{Un-Track}          & CVPR 2024 & ViT$-$B&$-$&$-$& $-$ & $-$& 84.2 & 62.5 & 66.7 & $-$ & 53.6 & $-$ \\
            & SDSTrack~\cite{SDSTrack}          & CVPR 2024 & ViT$-$B&$-$&$-$& $-$ & $-$& 84.8 & 62.5 & 66.5 & $-$ & 53.1 & 20.9 \\
            & IIMF~\cite{chen2024simplifying}   & ACM MM 2024 & ViT$-$B&$-$&$-$  & 85.6 &62.4 & 86.8 & 64.4&72.4 & 68.4 & 58.1 & $-$ \\
            & AINet~\cite{lu2024rgbt}   & AAAI 2025 & ViT$-$B&$-$&$-$  &\color{blue} 86.8 & \color{red}64.1& \color{blue}89.1 & \color{blue}66.8& \color{blue}73.0 &\color{blue} 69.0 & \color{blue}58.2 &\color{blue} 38.1 \\
            \midrule
            \multirow{5}{0.8cm}{\centering \small \raisebox{1cm}{\rotatebox{-90}{\textbf{\parbox{1.5cm}{Pixel$-$level\\ \hspace*{0.2cm}Fusion}}}}}
            & mfDiMP~\cite{zhang2019multi}      & ICCVW 2019 & ResNet$-$50& 83.6 & 69.7 & $-$ & $-$ & $-$ & $-$& 44.7 & 39.5 & 34.3 & 10.3 \\
            & ProTrack~\cite{ProRGBTTrack}      & ACM MM 2022 & ViT$-$B&$-$&$-$& $-$ & $-$& 78.6 & 58.7 & 50.9 & $-$ & 42.1 & 30 \\
            & X$-$Net~\cite{ding2023x}            & arXiv 2023  & VGG$-$M & \color{blue}93.1 & \color{red}76.7 & $-$ & $-$& 85.2 & 62.2 & 50.8 & $-$ & 44.6 & $-$ \\ 
            & DFAT~\cite{Exploring_IF_2023}     &IF 2023 &ResNet$-$50 & $-$ & $-$  & $-$ & $-$& 75.8 & 55.2& $-$ & $-$& $-$& 21 \\
            & \cellcolor{gray!30}TPF(Ours)           &\cellcolor{gray!30} $-$ &\cellcolor{gray!30} ViT$-$B&\cellcolor{gray!30}\color{red} 94.3&\cellcolor{gray!30}\color{blue} 76.3  &\cellcolor{gray!30}\color{red} 88.0 &\cellcolor{gray!30}\color{blue} 63.8 &\cellcolor{gray!30}\color{red} 89.7&\cellcolor{gray!30}\color{red} 67.1  &\cellcolor{gray!30} \color{red} 75.1 &\cellcolor{gray!30} \color{red} 71.3 &\cellcolor{gray!30} \color{red} 59.5&\cellcolor{gray!30} \color{red} 59.4\\
            \bottomrule
        \end{tabular}}
    
 \label{sota_compare_tab}
\end{table*}

\subsection{Implementation Details} We select OSTrack~\cite{ostrack} as the single-stream tracker and use ViT-B~\cite{ViT_network} as the feature extractor, and input resolution is 256 $\times$ 256. For parameter initialization, we leverage the powerful dropmae~\cite{wu2023dropmae} as the pre-trained model for the tracker. To initialize the pixel-level fusion adapter, we perform multi-expert distillation on the LasHeR training set. The adapter is then integrated with the tracker, and we fine-tune the combined model following the training setup of the tracking network.

During fine-tuning, the entire model utilizes AdamW~\cite{AdamW} to minimize task loss, reconstruction loss, and repulsion loss. We fine-tune the entire network in an end-to-end manner on the LasHeR training set and then directly evaluate it on four datasets: GTOT~\cite{li2016gtot}, RGBT210~\cite{Li17rgbt210}, RGBT234~\cite{li2019rgb234}, and LasHeR~\cite{li2021lasher}. 
In terms of learning rates, we set the backbone network of the fusion branch and the pixel-level fusion adapter to 5e-6, while the tracking head learning rate is set to 5e-5. Additionally, the learning rates for the RGB and TIR branches are both set to 5e-5. The model is implemented on the PyTorch platform, using an Nvidia RTX 4090 GPU with 24 GB of memory and a batch size of 18. The distillation and full fine-tuning takes 35 and 30 epochs, respectively, with each epoch consisting of 60,000 sample pairs.
During the inference phase, we set the update interval $N$ to 50 for the LasHeR, RGBT234, and RGBT210 datasets, and 15 for the GTOT dataset due to its shorter sequences, and set the reliability threshold $p$ to 0.65.

\subsection{Evaluation Dataset and Protocol}

\noindent\textbf{Dataset.} We evaluate our model on multiple RGBT tracking datasets to ensure a comprehensive assessment. The earliest dataset, \textbf{GTOT}, consists of 50 sequences with a total of approximately 15,000 frames. However, its average sequence length of only 150 frames constrains the scope of evaluation.
The \textbf{RGBT210} dataset extends this with 210 sequences, encompassing 209.4K frames, while \textbf{RGBT234} provides further refinements, offering improved data alignment and more precise bounding boxes and annotations across 12 challenging attributes, totaling 233.4K frames. 
The largest dataset, \textbf{LasHeR}, comprises 1,224 video sequences (1469.6K frames) and is divided into 245 test and 979 training sequences, offering a comprehensive benchmark for performance evaluation in RGBT tracking.

\noindent\textbf{Protocol.} For evaluation, we adopt precision rate (PR) and success rate (SR) as key metrics under the one-pass evaluation (OPE) protocol, both widely used in RGBT tracking. \textbf{PR} measures the proportion of frames where the distance between the tracker’s output and the true bounding box is below a threshold (5 pixels for GTOT, 20 for others). \textbf{SR} calculates the proportion of frames with an Intersection over Union (IoU) above a threshold, with the final SR score obtained by varying this threshold to compute the Area Under the Curve (AUC). The normalized PR (\textbf{NPR}) adjusts the PR by the scale of the ground truth box, with a representative NPR score computed by calculating the area under the curve for normalized thresholds within [0, 0.5].

\subsection{Comparison with State-of-the-art Methods}

We compare our method, TPF, with 27 state-of-the-art RGBT trackers across four major benchmarks, including LasHeR~\cite{li2021lasher}, RGBT234~\cite{li2019rgb234}, RGBT210~\cite{Li17rgbt210}, and GTOT~\cite{li2016gtot}.
As shown in Table~\ref{sota_compare_tab}, TPF achieves the highest performance on the LasHeR dataset, with PR, NPR, and SR scores of 75.1\%, 71.3\%, and 59.5\%, respectively. This surpasses both feature-level and pixel-level fusion trackers, including the leading feature-level tracker, AINet. Specifically, TPF outperforms AINet by 2.1\%, 2.3\%, and 1.3\% in PR, NPR, and SR metrics, demonstrating its obvious performance advantages.
On the RGBT234 dataset, TPF also exceeds existing feature-level fusion and pixel-level fusion trackers, achieving PR and SR scores of 89.7\% and 67.1\%. Compared to AINet, the second-best method on this dataset, our approach demonstrates a clear advantage, exceeding AINet by 0.6\% in PR and 0.3\% in SR.
Similarly, on the RGBT210 and GTOT datasets, TPF maintains outstanding performance, further illustrating its robustness and generalization ability across various datasets, including smaller ones.
These results highlight the effectiveness and robustness of our pixel-level fusion method, making it promising as a strong alternative to the current mainstream feature-level fusion methods for RGBT tracking, due to its high performance and gap-free properties.

\subsection{Ablation Studies} 
To validate the effectiveness of the proposed method, we conduct a series of ablation studies, including quantitative and qualitative experiments, on the currently largest RGBT tracking dataset, LasHeR~\cite{li2021lasher}.

\begin{table}[h]
\centering
\caption{Ablation study of the proposed TPF network.}
\renewcommand\arraystretch{1}
\resizebox{0.45\textwidth}{!}{
\begin{tabular}{c|c|c|c|c}
\toprule
\textbf{Method}                      & \textbf{Precision} & \textbf{NormPrec} & \textbf{Success} & \textbf{FPS} \\ \hline \hline
Baseline            & 67.0      & 62.9     & 53.3 & 99.8    \\ \hline
w/ PFA              & 68.9          & 65.2         &  54.8  & 64.1     \\
w/ PFA TPL     & 72.5      & 68.5     & 57.7   & 64.1\\ 
w/ PFA NDTU          & 72.1      & 68.0    & 57.1    & 59.4  \\
\rowcolor{gray!30} w/ PFA TPL NDTU & \textbf{75.1}      & \textbf{71.3}     & \textbf{59.5}  & 59.4 \\ 

\bottomrule
\end{tabular}}

\label{tab:Component_ablation}
\end{table}

\noindent\textbf{Component Analysis.}
In Table~\ref{tab:Component_ablation}, we conduct ablation studies to verify the effectiveness of different modules in TPF. To fairly validate the effectiveness of the proposed components, we construct several variants as follows.

\textit{Baseline} denotes a single-stream tracker that performs pixel-level fusion of RGB and TIR images using addition, which is also used in existing pixel-level fusion effort~\cite{ProRGBTTrack}. 

\textit{w/ PFA} denotes the baseline model equipped with pixel-level fusion adapter (PFA) and then the directly fine-tuned with tracking task, which achieves a certain improvement, surpassing the \textit{Baseline} by \textbf{1.9\%/2.3\%/1.5\%} in PR/NPR/SR metrics. The experiment shows that the structure of PFA is effective, but still falls short of existing feature-level fusion methods.

\textit{w/ PFA TPL} denotes the addition of a task-driven progressive learning (TPL) framework to \textit{w/ PFA}. This framework first leverages multi-expert adaptive distillation to pre-train \textit{w/ PFA} and then applies decoupling representation fine-tuning in all model parameters. The TPL approach yields substantial performance gains, with PR/NPR/SR improvements of \textbf{3.6\%/3.2\%/2.9\%} over \textit{w/ PFA}.

\textit{w/ PFA NDTU} denotes the addition of a nearest-neighbor dynamic template updating (NDTU) strategy to \textit{w/ PFA}. The NDTU strategy also brings significant performance improvements, with PR/NPR/SR metrics increased by \textbf{3.2\%2.8\%2.3\%} over \textit{w/ PFA}.

\textit{w/ PFA TPL NDTU} denotes the \textit{w/ PFA TPL} equipped with NDTU strategy, resulting in notable performance gains of \textbf{2.6\%/2.8\%/1.8\%} over \textit{w/ PFA TPL}. 
%
%
These results demonstrate the effectiveness of the NDTU strategy.

\noindent\textbf{Impact of Distillation and Fine-tune.} To evaluate the contributions of the key components, multi-expert distillation (MAD) and decoupling representation fine-tuning (DRF), in the TPL framework, we present ablation results in Table~\ref{tab:TPL_ab}. The results show that removing either DRF or MAD leads to clear performance degradation, indicating that both distillation and fine-tuning are essential for enhancing the performance of pixel-level fusion tracking.

\begin{table}[h]
\centering
\caption{Ablation study of the TPL.}
\renewcommand\arraystretch{1}
\resizebox{0.45\textwidth}{!}{
\begin{tabular}{c|cc|c|c|c}
\toprule
\multirow{2}{*}{\textbf{Variants}} & \multicolumn{2}{c|}{\textbf{TPL}}                     & \multirow{2}{*}{\textbf{Precision}} & \multirow{2}{*}{\textbf{NormPrec}} & \multirow{2}{*}{\textbf{Success}} \\ 
 & \textbf{MAD}              & \textbf{DRF}   &       &       &    \\ \hline \hline
\textbf{1}                         & \checkmark &                           & 70.8                         & 66.9                          & 56.3                         \\
\textbf{2}                         &                           & \checkmark & 71.1                         & 67.0                          & 56.1                         \\
\rowcolor{gray!30}\textbf{3}                         & \checkmark & \checkmark & 72.5                         & 68.5                          & 57.7                         \\ \bottomrule
\end{tabular}}

\label{tab:TPL_ab}
\end{table}

\noindent\textbf{Impact of Reconstruction and Repulsion Losses.} To further validate the effectiveness of the decoupling representation fine-tuning (DRF) strategy, we conduct an ablation study on the two associated losses, as shown in Table~\ref{tab:DRF_ab}. The results indicate that removing the reconstruction loss results in a performance drop, and removing the repulsion loss causes a further decline, demonstrating the critical role of both losses in the DRF.

\begin{table}[h]
\centering
\caption{Ablation study of the DRF.}
\renewcommand\arraystretch{1}
\resizebox{0.45\textwidth}{!}{
\begin{tabular}{c|cc|c|c|c}
\toprule
\multirow{2}{*}{\textbf{Variants}} & \multicolumn{2}{c|}{\textbf{DRF}}                     & \multirow{2}{*}{\textbf{Precision}} & \multirow{2}{*}{\textbf{NormPrec}} & \multirow{2}{*}{\textbf{Success}} \\ 
 & \textbf{$\mathcal{L}_{rep}$}              & \textbf{$\mathcal{L}_{rec}$}   &       &       &    \\ \hline \hline
\textbf{1}                         &  &                           & 70.8                         & 66.9                          & 56.3                         \\
\textbf{2}                         & \checkmark       &  & 71.8                        & 67.9                          & 57.2                      \\
\rowcolor{gray!30}\textbf{3}                         & \checkmark & \checkmark & 72.5                         & 68.5                          & 57.7                         \\ \bottomrule
\end{tabular}}
\label{tab:DRF_ab}
\end{table}

\noindent\textbf{Impact of Model Complexity.} To assess the effect of model complexity, we analyze the relationship between parameter count and performance, as shown in Table~\ref{tab:model_complexity}. Integrating PFA slightly increases parameters (92.136M vs. 92.122M) but provides only a marginal improvement (68.9 PR vs. 67.0 baseline). However, introducing MAD and DRF, which enhance fusion during training, significantly boosts performance (70.8 PR and 71.1 PR, reaching 72.5 PR when combined). Importantly, these improvements occur without increasing test-time parameters (92.136M), confirming that the performance gains stem from more effective fusion rather than increased model complexity.
\begin{table}[h]
\centering
\caption{Ablation study of model complexity.}
\renewcommand\arraystretch{1.5}
\resizebox{0.48\textwidth}{!}{
\begin{tabular}{l|c|c|c|c|c}
\toprule
\textbf{Method}  & \textbf{Precision}  & \textbf{NormPrec} & \textbf{Success}  & \textbf{Train Params} & \textbf{Test Params} \\ \hline 
\midrule
Baseline          & 67.0  & 62.9  & 53.3  & 92.122M  & 92.122M  \\ 
w/ PFA           & 68.9  & 63.2  & 54.8  & 92.136M  & 92.136M  \\ 
w/ PFA MAD     & 70.8  & 66.9  & 56.3  & 92.136M  & 92.136M  \\ 
w/ PFA DRF     & 71.1  & 67.0  & 56.1  & 290.201M & 92.136M  \\ 
w/ PFA MAD DRF & 72.5  & 68.5  & 57.7  & 290.201M & 92.136M  \\ 
\bottomrule
\end{tabular}}

\label{tab:model_complexity}
\end{table}

\noindent\textbf{Impact of Multi-expert.} To evaluate the contributions of the multi-expert design, we present the results of single-expert distillation in Table~\ref{tab:MAD_ab}. Each expert is used to distill the PFA individually and then directly fine-tuned with the tracker. The results indicate that single-expert distillation yields lower performance than multi-expert distillation, highlighting the importance of the multi-expert design.
\begin{table}[h]
\centering
\caption{Ablation study of the MAD.}
\renewcommand\arraystretch{1}
\resizebox{0.45\textwidth}{!}{
\begin{tabular}{c|c|c|c}
\toprule
\textbf{Method}  & \textbf{Precision} & \textbf{NormPrec} & \textbf{Success} \\ \hline \hline
w/ PFA CDDFuse$_{D}$       & 69.7      & 65.7     & 55.5  \\
w/ PFA Diff-IF$_{D}$              & 70.3          & 66.5         &  56.1       \\
w/ PFA MambaDFuse$_{D}$              & 69.9          & 66.0         &  55.7       \\ \hline
\rowcolor{gray!30} w/ PFA MAD           &  70.8         & 66.9         & 56.3     \\  
\bottomrule
\end{tabular}}

\label{tab:MAD_ab}
\end{table}


\noindent\textbf{Advantages of PFA.} To demonstrate the advantages of the pixel-level fusion adapter (PFA), we compare its performance against various image fusion methods integrated with the tracker, as shown in Table~\ref{tab:PFA_ab}. Since fine-tuning these fusion models for tracking is computationally expensive, we adopt a consistent approach by using their fused images to train and test the tracker. The results show that PFA* achieves comparable or superior performance with fewer parameters, lower GFLOPs, and faster speed than state-of-the-art fusion models, highlighting the efficiency and effectiveness of our design. Notably, since the pixel-level fusion scheme of this paper is driven by a tracking task, we only provide the performance of PFA on image fusion data in the \textbf{Supplementary Material}.
\begin{table}[h]
\caption{Performance comparison of advanced image fusion models combined with a tracker. PFA* denotes the frozen parameters.}
\renewcommand\arraystretch{1}
\resizebox{0.48\textwidth}{!}{
\begin{tabular}{c|c|c|c|c|c}
\toprule
\textbf{Method}                      & \textbf{Precision} & \textbf{Success} & \textbf{Param} & \textbf{FLOPs} & \textbf{FPS}\\ \hline \hline
w/ CDDFuse~\cite{zhao2023cddfuse}                       &  63.5         & 50.9     & 1.19M  & 117.04G  & 38.9  \\
w/ Diff-IF~\cite{yi2024diff}          &  61.5               & 48.2    & 23.74M  & 696.48G  & 4.7    \\
w/ MambaDFuse~\cite{li2024mambadfuse}      & 62.2         & 49.9  & 1.35M  & 12.35G  & 22.3  \\
\rowcolor{gray!30} w/ PFA* & 64.6          & 51.4   & 0.014M  & 0.283G  & 201.8  \\
\bottomrule
\end{tabular}}

\label{tab:PFA_ab}
\end{table}

\subsection{Analysis and Visualization}


\noindent\textbf{Challenging-based Performance.} We analyze the performance of TPF in various challenging scenarios by evaluating it on the LasHeR dataset under different challenging subsets. Due to space limitations, detailed experiments are provided in the supplementary material.
%

\begin{figure}[h]
	\centering
	\includegraphics[width=0.8\linewidth]{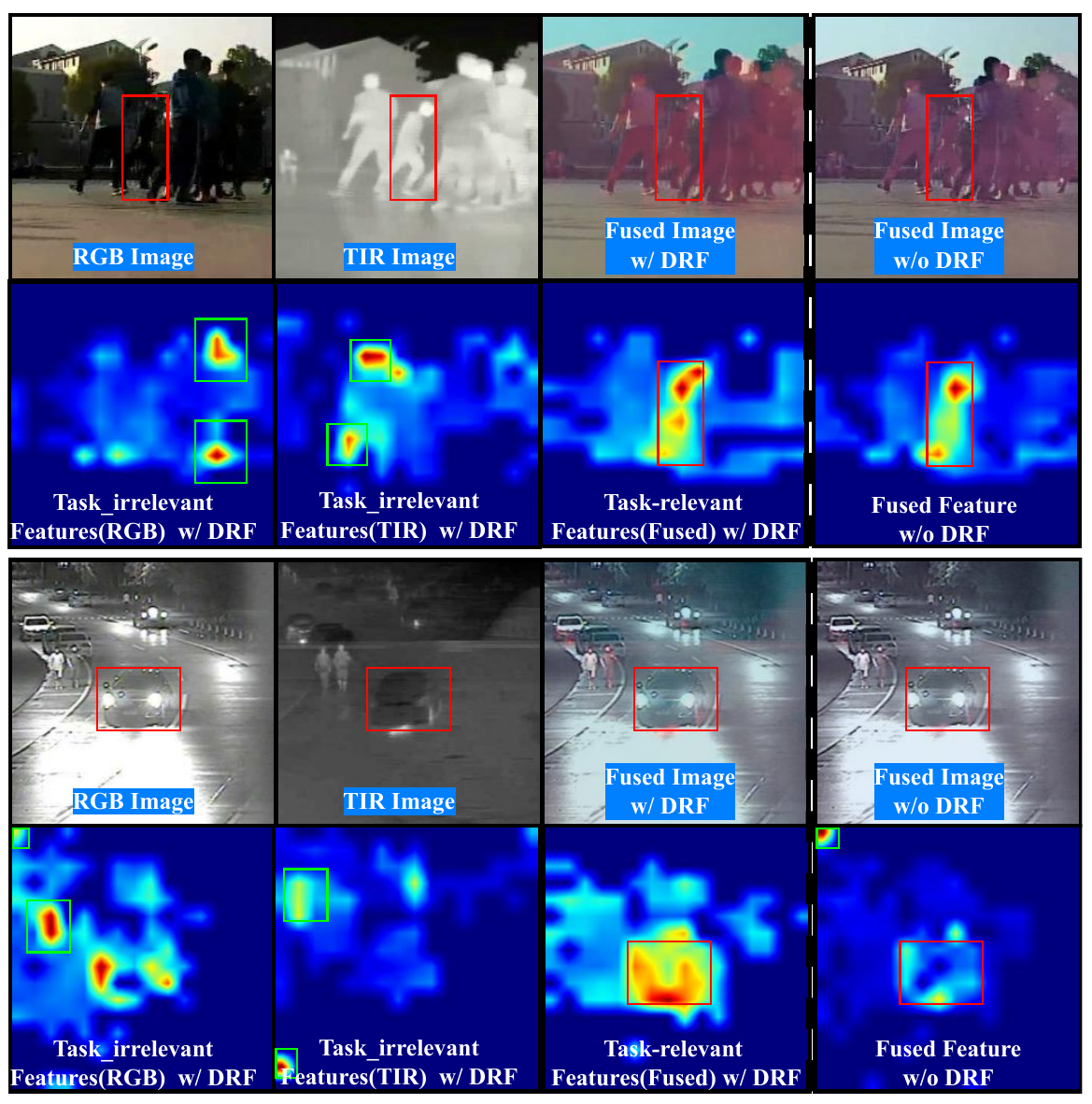}
	\caption{
	Visualization of fused images, fused features, and decoupled features with and without DRF. Green boxes indicate task-irrelevant features, while red boxes highlight task-relevant features.
	}
	\label{fig:feature_vis}
\end{figure}

\noindent\textbf{Visualization.} To better understand the fused images, fused features, and decoupled features, we visualize the fused image and fused features between ours approach with decoupled representation fine-tuning (i.e., w/ DRF)  and ours approach without the DRF (i.e., w/o DRF) in Figure~\ref{fig:feature_vis}. It can be observed that the fused images successfully retain the beneficial information from both modalities while minimizing low-quality information, as shown in row one and three. While the differences in the fused images before and after introducing DRF may not be easily noticeable to the human vision, we observe that the incorporation of DRF in the fused features enhances the model's focus on the target, as demonstrated in rows two and four. Finally, we also visualize the feature maps from the RGB and TIR branches in w/ DRF, which primarily focus on non-target areas, i.e., task-irrelevant information. These observations highlight the effectiveness of explicitly decoupling task-relevant and task-irrelevant learning for pixel-level fusion tracking.
In addition, we also provide more visualization results in \textbf{Supplementary Materials}, including fusion image visualization, tracking result visualization, etc.

\section{Conclusion}

In this paper, we propose the Task-driven Pixel-level Fusion (TPF) network to break the limitations of shallow interactions and fully exploit the gap-free advantage in pixel-level fusion RGBT tracker.
By introducing a lightweight pixel-level fusion adapter (PFA) for efficient fusion and a task-driven progressive learning framework the discriminative power of shallow network, TPF significantly enhances the performance of pixel-level fusion tracking. Additionally, we introduce a nearest-neighbor dynamic template updating strategy to handle appearance changes during tracking. Experimental results show that TPF outperforms most existing trackers on multiple RGBT tracking datasets, achieving impressive performance with lower computational overhead.

\textbf{Limitation.} Although PFA significantly reduces the number of parameters and computational demands, its inference speed is limited by the lack of adaptation to current acceleration hardware. To address this, we plan to implement the newly developed Mamba2~\cite{dao2024transformers}, aiming to achieve high-speed inference.
Moreover, while our fusion design is lightweight and efficient, it remains a serial architecture that operates independently of the tracker, which introduces some computational overhead. In future work, we could explore a unified architecture that integrates pixel-level fusion directly with tracking, enabling both processes to be handled within a single branch.
%




\section{Supplementary Material}


In the supplementary materials, we present three experiments to comprehensively demonstrate the advantages and robustness of TPF. These include comparisons with image fusion methods, parameter sensitivity analysis, challenging-based performance evaluation and tracking results visualization.



\subsection{Comparison with image fusion methods}

\begin{figure*}[ht]
	\centering
	\includegraphics[width=0.9\linewidth]{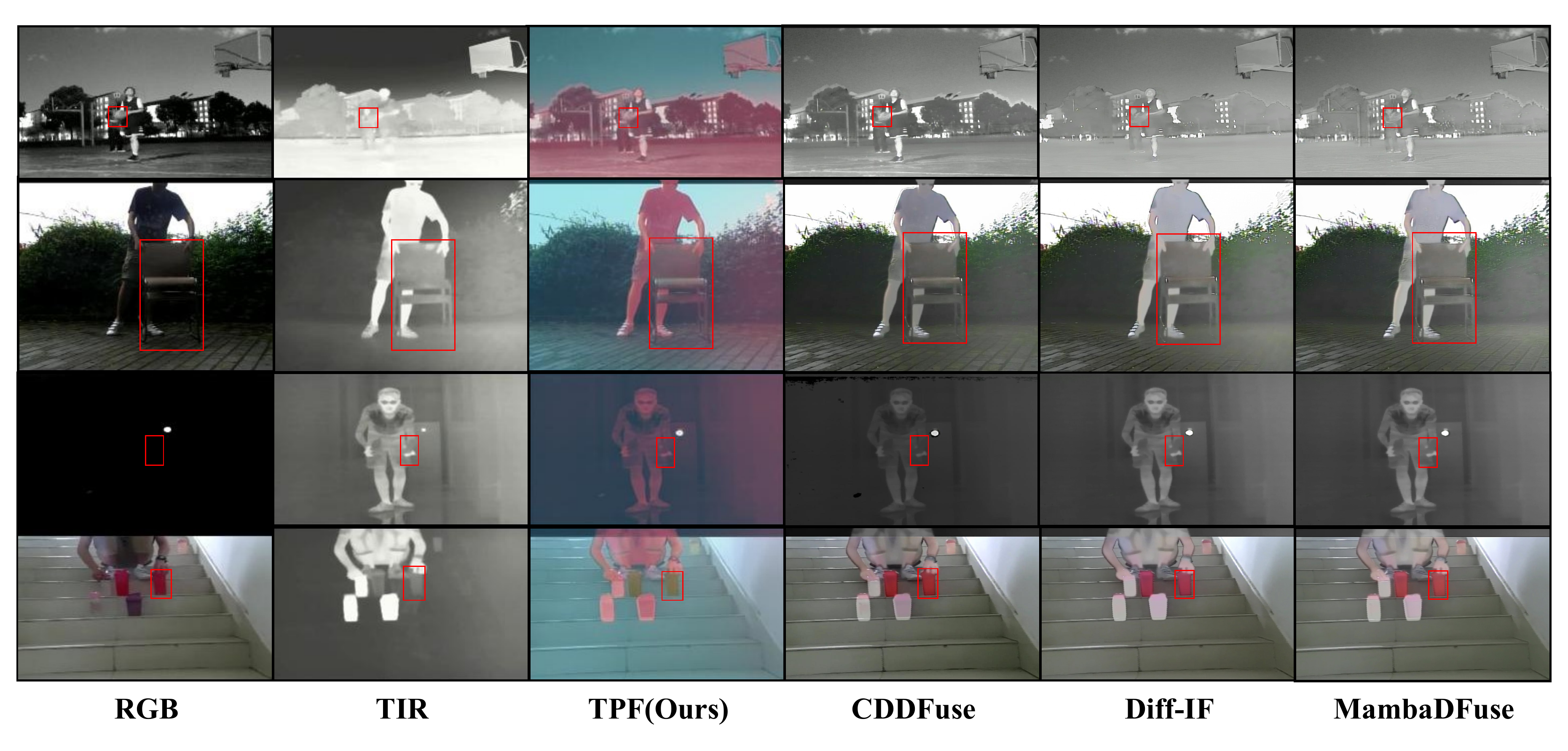}
	\caption{
	Comparison of fusion results between the pixel-level fusion method and the other three image fusion methods on the LasHeR dataset.
	}
	\label{fig:fused_image}
\end{figure*}

To demonstrate the difference between pixel-level fusion oriented to tracking and image fusion oriented to human vision, we visualize the difference between TPF and three existing image fusion methods (CDDFuse~\cite{zhao2023cddfuse}, Diff-IF~\cite{yi2024diff}, and MambaDFuse~\cite{li2024mambadfuse}) by presenting the results of fused images on the LasHeR~\cite{li2021lasher} dataset, as shown in Fig.~\ref{fig:fused_image}.
Compared to the other three methods, the fusion results of our approach in the tracking target region demonstrate superior saliency and information completeness. For instance, in the first and second rows, our method clearly enhances the visibility of the target region with higher saliency. In contrast, the third and fourth rows reveal a distinct difference between the performance of pixel-level fusion and existing image fusion techniques. Specifically, in image fusion, non-salient target information from a particular modality is often discarded. However, pixel-level fusion effectively integrates information from both modalities, ensuring comprehensive fusion. In other words, pixel fusion fully leverages modal information that may be less significant in the target, which could be crucial for tracking tasks. In contrast, existing image fusion techniques primarily emphasize visual brightness, often overlooking the specific requirements of machine vision.

In addition, we also explore impact of tracking task on image fusion task. We compared our image fusion performance with existing methods on the MSRS dataset~\cite{tang2022piafusion1} under the same conditions. The results are as follows:
\begin{table}[h]
\small
\centering
\resizebox{0.48\textwidth}{!}{
\renewcommand{\arraystretch}{0.8}
\begin{tabular}{lp{0.3cm}p{0.3cm}p{0.3cm}p{0.6cm}p{0.8cm}p{0.8cm}}
\toprule
\textbf{Method} & \textbf{EN$\uparrow$}  & \textbf{SF$\uparrow$} & \textbf{MI$\uparrow$}& \textbf{MSE$\uparrow$} & \textbf{PSNR$\uparrow$} & \textbf{SSIM$\uparrow$} \\
\hline
\midrule
CDDFuse & 6.70  & 5.86 & 3.29& 50.52&31.23 &1.39\\
Diff-IF & 6.68  & 5.93 & 3.37&46.44 &31.53 &1.39 \\
PFA  & 6.63  & 5.68 & 2.39 &58.71 &30.48 &1.35\\
PFA w/ tracking & 6.17  & 5.18 & 2.32&54.23 &26.38 &0.92 \\
\bottomrule
\end{tabular}}
\end{table}

The fusion metrics of PFA decreased with the introduction of tracking tasks, which allows us to infer that the two task objectives are different. Tracking focuses only on the target area and does not help with the global image fusion task, it is possible that segmentation tasks could help enhance image fusion tasks.

\subsection{Parameter sensitivity analysis.}
\begin{figure}[h]
	\centering
	\includegraphics[width=0.85\linewidth]{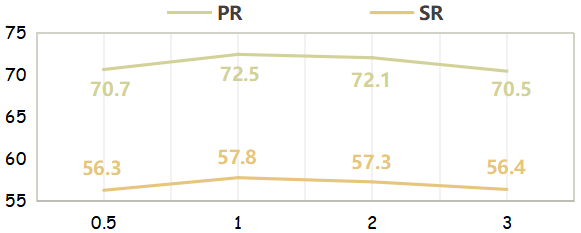}
	\caption{
	Sensitivity analysis of the loss weight $\lambda$, with the horizontal axis representing different $\lambda$ values and the vertical axis showing the corresponding evaluation scores.
	}
	\label{fig:parameter_ablation}
\end{figure}
we analyze the sensitivity of our model to different loss weight $\lambda$ for the decoupled representation fine-tuning loss and the task loss in Figure~\ref{fig:parameter_ablation}. It can be seen that there is a clear degradation in performance when we over-weaken and over-strengthen the effects of the DRF, while the model performs better when the weights of the two losses are relatively balanced and achieves the best performance when $\lambda$ is set to 1.

\subsection{Challenging-based performance evaluation}

\begin{figure*}[h]
	\centering
	\includegraphics[width=0.9\linewidth]{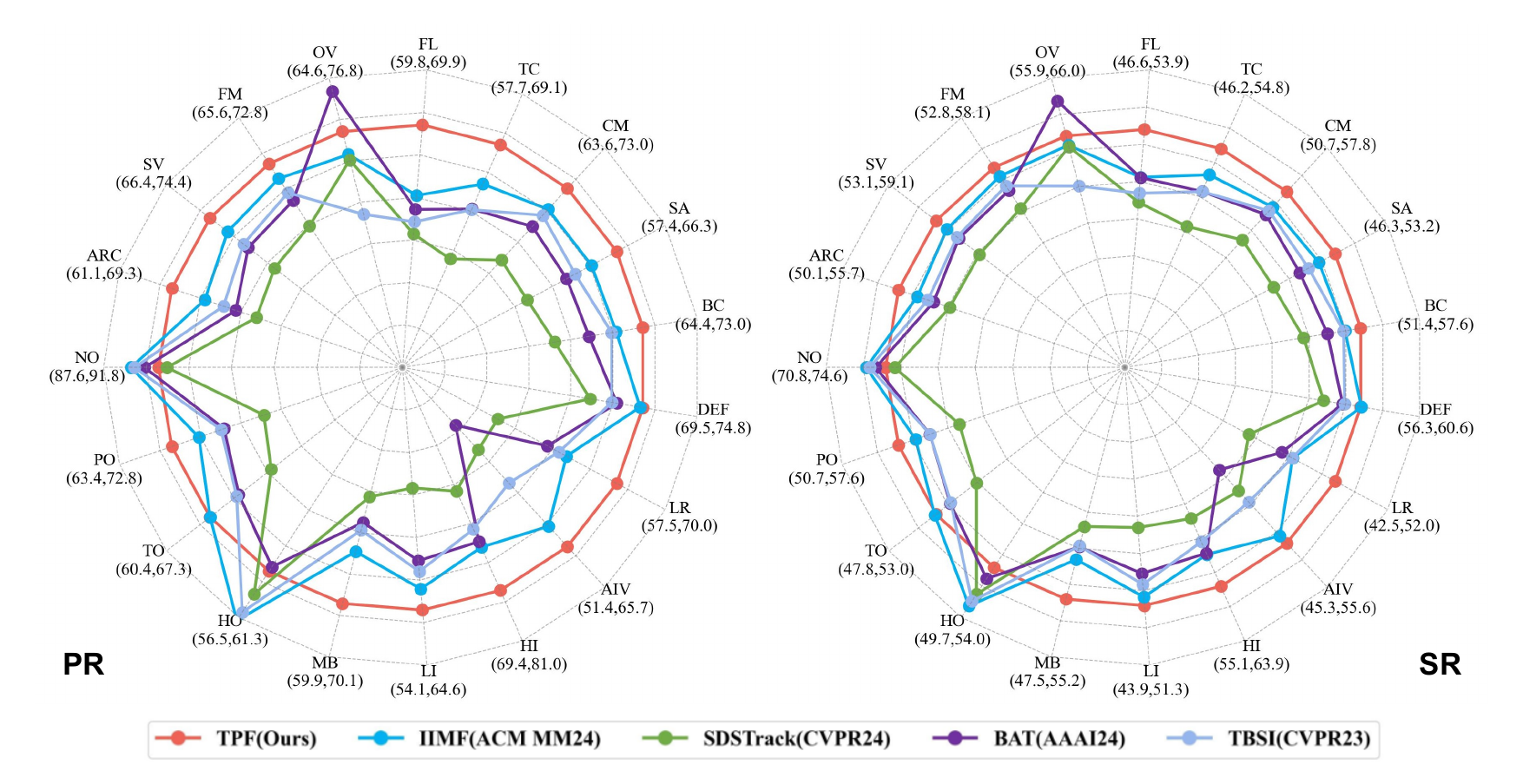}
	\caption{
	Challenge-based performance evaluation on the LasHeR dataset. The left side displays challenge-based precision rate, while the right side
shows challenge-based success rate. The axes of each attribute have been normalized. For clarity, only the minimum and maximum values for each attribute metric are presented. 
	}
	\label{fig:LaSheR_PR_SR}
\end{figure*}

To evaluate the robustness of TPF in various challenging tracking scenarios, we compare our method with several advanced feature-level trackers, including IIMF~\cite{chen2024simplifying}, SDSTrack~\cite{SDSTrack}, BAT~\cite{BAT2024}, and TBSI~\cite{TBSI}, on the LasHeR~\cite{li2021lasher} dataset. The dataset includes 19 distinct challenges: No Occlusion (NO), Partial Occlusion (PO), Total Occlusion (TO), Hyaline Occlusion (HO), Out-of-View (OV), Low Illumination (LI), High Illumination (HI), Abrupt Illumination Variation (AIV), Low Resolution (LR), Deformation (DEF), Background Clutter (BC), Similar Appearance (SA), Thermal Crossover (TC), Motion Blur (MB), Camera Moving (CM), Frame Lost (FL), Fast Motion (FM), Scale Variation (SV), and Aspect Ratio Change (ARC). As shown in Fig.~\ref{fig:LaSheR_PR_SR}, TPF outperforms these trackers across most challenges, achieving excellent results in all tested scenarios.
%

In most challenges, TPF based on pixel-level fusion demonstrates a clear advantage, particularly in challenges such as FL, TC, MB, FM, and LR, highlighting the effectiveness of our method. Specifically, in challenges like FL and TC, the fused images generated by pixel-level fusion help avoid issues such as frame loss or instability in tracking continuity caused by thermal crossover. In MB and FM scenarios, pixel-level fusion mitigates the imaging sensitivity of the RGB modality under these challenges. In LR, pixel-level fusion better utilizes low-level detail information. \textit{This shows that pixel-level fusion significant advantage in handling challenges posed by data imaging.}
However, in certain challenges, such as NO, HO, and OV, pixel-level fusion shows some limitations compared to feature-level fusion. These challenges often stem from the complexity of the target's motion. \textit{This suggests that feature-level fusion is better suited for handling complex motion scenarios. Therefore, this observation suggests that there is a complementarity between these two types of methods in different challenge scenarios.}



\subsection{Tracking results visualization}

\begin{figure*}[ht]
	\centering
	\includegraphics[width=0.9\linewidth]{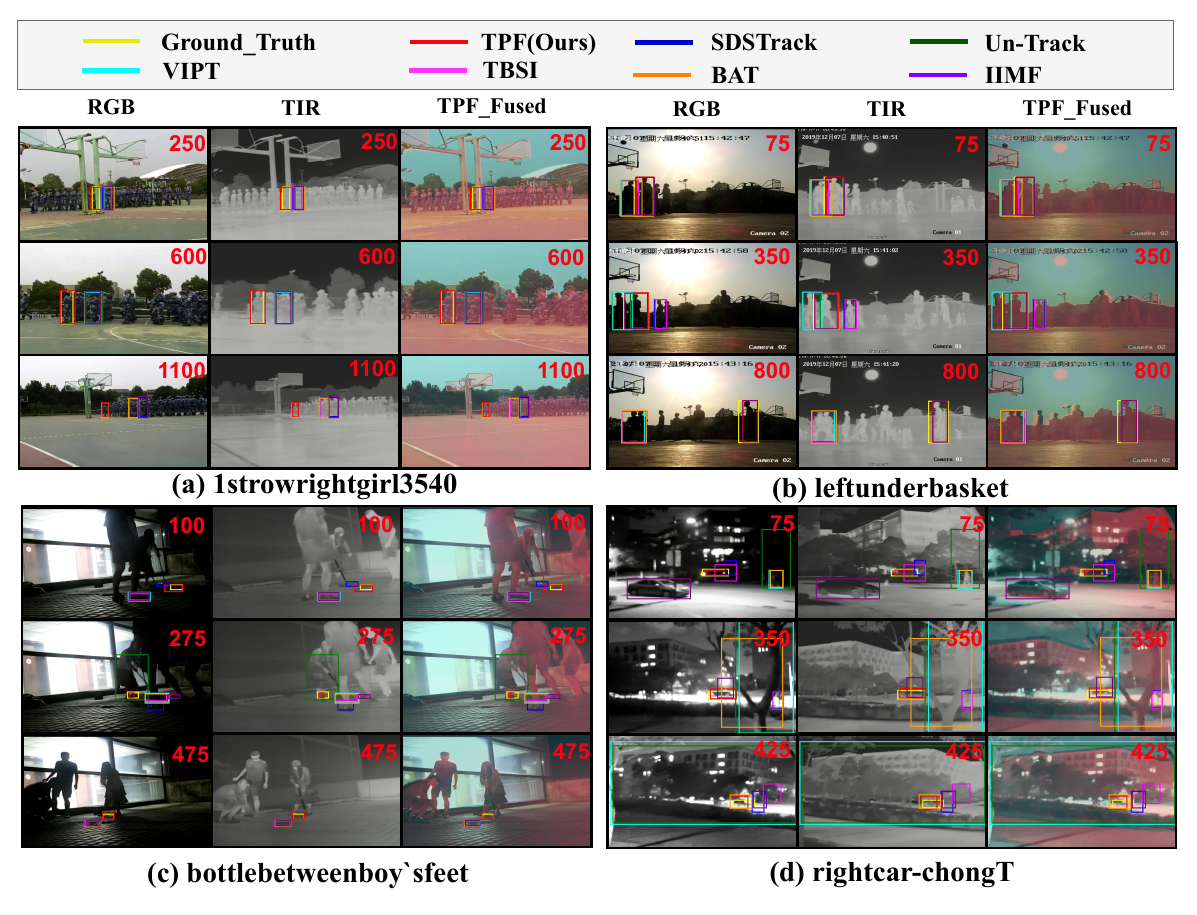}
	\caption{
		Qualitative comparison of TPF against six state-of-the-art feature-level trackers on four representative video sequences.
	}
	\label{fig:box_image}
\end{figure*}

To evaluate the effectiveness and robustness of our method, we compare it with six state-of-the-art trackers, including SDSTrack~\cite{SDSTrack}, Un-Track~\cite{Un-Track}, VIPT~\cite{ViPT}, TBSI~\cite{TBSI}, BAT~\cite{BAT2024}, and IIMF~\cite{chen2024simplifying}, on the LasHeR~\cite{li2021lasher} dataset, as illustrated in Fig.~\ref{fig:box_image}. 
Specifically, in sequence (a), during the total occlusion (TO) challenge at frame 250, where the target is fully occluded by the basketball hoop, our method accurately tracks the target without losing its position. At frames 600 and 1100, significant distractions with similar appearances (SA) cause other trackers to fail, while ours maintains precise localization. In sequence (b), our method excels in high illumination (HI) and similar appearance (SA) challenges, consistently tracking the target over extended intervals. For low illumination (LI) challenges, especially in sequence (d), where two trackers fail completely, our method achieves significantly higher precision. 
\textit{In conclusion, through visual analysis of TPF with six advanced feature-level trackers in these representative sequences, the robustness of our method in challenging tracking scenarios is clearly demonstrated, highlighting the feasibility of pixel-level tracking.}
\bibliographystyle{IEEEtran}
\bibliography{main} 

\end{document}